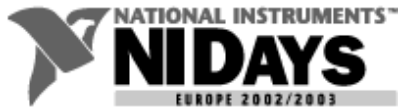

# DISTRIBUTED AND PARALLEL NET IMAGING

**Gerardo Iovane**
Researcher
**Dipartimento di Ingegneria Informatica e Matematica Applicata – Università degli Studi di Salerno**

Research, Imaging Equipment, University/Education

**Products used**
- LABVIEW  Prof Dev Sys 6.1
- LABVIEW  Real Time Module 6.1
- IMAQ Vision 6.01
- SQL TOOLKIT 2.0
- INTERNET TOOLKIT 5.0
- SIGNAL PROCESSING TOOLSET
- SPC Tools

**The challenge:** The development of a distributed and parallel net imaging system in order to perform very fast and real time image acquisition, reduction and analysis by very large machine vision with the biggest images (even 256Mpixels).
.

**The solution:** A real time network parallel image processing system is developed using Labview. The architecture is provided with a server, linked with a maximum of 256 workstations to analyse images with a size of 256 Mpixels. The server takes data directly from the acquisition system, which is formed by a telescope with 16k×16k pixels CCD camera.

## Abstract

A very complex  vision system is developed to detect luminosity variations connected with the discovery of new planets in the Universe. The traditional imaging system can not manage a so large load. A private net is implemented to perform an automatic vision and decision architecture. It lets to carry out an on-line discrimination of interesting events by using two levels of triggers. This system can even manage many Tbytes of data per day. The architecture avails itself of a distributed parallel network system based on a maximum of 256 standard workstations with Microsoft Window as OS.

## Introduction
The passage of the planet close to the line of sight of the observer implies a luminosity variation of the source star in the image. This is a rare event; there are many difficulties due to the large amount of monitored luminous objects, corresponding to never realized CCD camera (256 Mpixels per image). A parallel system, distributed in the world is an effective solution. It can analyse data and pre-processed images in real time.
From a conceptual point of view the steps to process the data may be summarized as follows:
1. Image data acquisition (CCD camera and grabber are realized ad hoc);
2. Technical Image Processing (pre-reduction for bias, dark, pixel to pixel variations, geometrical and photometric calibration);
3. First Trigger Level (selection of interesting luminosity variations trough a peak detection algorithm);

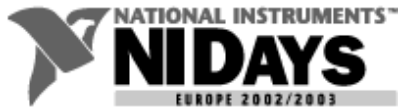

4. Second Trigger Level (Selection of Events compatible with the models);
5. Pilot Analysis;
6. Off - line Analysis.

The first two points are performed by the server central unit, while the others by the client systems.

## Environment

The system architecture is structured as shown in Fig.1. All software is realized with Labview and Labview RT. In add the image analysis is realized with IMAQ and the data analysis use the SIGNAL PROCESSING TOOLSET by NI.

**The server machine controls the following units** (see control panel in Fig.2)**.**

**i)** The Image Data Acquisition (I-DAQ) Unit is responsible for the data acquisition and pre-reduction of data coming from images.

**ii)** The Control Unit, which thanks to Telescope Control System (TCS), controls the telescope by following the instructions provided by the DataBase Control System (DBCS) or by the user.

**iii)** The DataBase (DB) Unit performs the data storage and processing according to simulations or previous observations. The DB use SQL language, and is linked to the system by SQL TOOLKIT by NI. Then each image will be divided into 256 maps of 1024×1024 pixels.

These maps are given to the 256 client machines, which perform the following operations and then give back the results to the Image Data Mining Unit. This unit rebuild the image on the original size (16k×16k pixels), but the original 250 Mbytes are reduced to 1-3 Mbytes, trough an high compression factor.

**The clients contains the following units** (see control panel in Fig.3)**.**

**iv)** The Processing and Analyzing (P&A) Unit is the platform where massive data analysis is made. It consists of two main units: 1) Data Processing Unit (DAP) for peak detection of relevant luminosity variation, and to remove usefulness objects; 2) Data Analysis Unit (DAU) for the fits of light curve with different expected models, color correlation, $\chi^2$ test. The DAP Unit is the first trigger level and is formed by four section: a) the peak detection procedure, b) the star (extended object) detection and filtering algorithm, c) the cosmic rays filter (rejection of saturated pixel), and d) the peak classification (single, double, multiple peak curve) methods. It is relevant to stress that for this step it can be used either the standard automatic procedures (where the peak detection method by NI is adopted) or the automatic unsupervised fuzzy neural network procedure (that is developed "ad hoc" with three different learning method: Multi-Layer-Self-Organizing-Maps, Neural gas, Maximum entropy). The second trigger level tests whether the measured luminosity variations are compatible with fixed and well known models or not, trough the Levemberg-Marquardt fit method implemented by NI.

**v)** The last Unit, so called Dispatcher, automatically builds status report about the different phases of the data flow starting from the I-DAQ to the DAU unit. Statistics, plots of data and events are produced and stored by this module. Moreover, in the occurrence of a special events, such as an alert, this unit can automatically reach people with an e-mail service and SMS (Short Message System). It is full developed by using the INTERNET TOOLKIT and the SPC tools by NI.

## Results of the software modules

In Fig.4 is shown a typical field of view, while in Fig.5 we can see the effect of the photometric alignment on the spectrum. Thanks to a filtering in the Fourier transformed space (see Fig.6) all "noised" objects will be removed (see Fig.7). Then, the light curves are built for each remaining pixel (see Figs.8).

## Conclusion

The architecture is a high tech challenge designed for very large machine vision and to process many Tbytes of data per day for new automated planet searches. This is performed with a network distributed parallel computing system. Most of the tools presented in this paper can be applied in other fields, where the image mining procedures are called for very large amount of data. The real time system is particularly useful to automatically select and to study light variation according to a fixed model. In terms of time load these procedures can produce results that can not reach with any other traditional approach. Moreover, thanks to the real time light curves monitoring and to the dispatcher implementation, the detection of short events and events with a huge main peak with a secondary one near the first (like in planetary system) become reachable. This is an other fundamental shoot of our system. Thanks to the presented system the Scientific Community can counts many interesting events, which are compatible with discovery of new planets outside the Solar System.

## Acknowledgements

The author wish to thank A.Ambu, I.Piacentini, M.Quaglia, F.Selleri, from NI Italy for relevant software suggestions and comments.

## References

1) G.Iovane, S.Capozziello, G.Longo, MEDEA: a real time imaging pipeline for pixel lensing, accepted by New Astronomy, 2002.
2) G.Iovane, MEDEA: Automated Measure and on-line Analysis in Astronomy and Astrophysics for Very Large Vision Machine, Proceeding of NIDays 2002, pag.171, VNU Business Publications, National Instruments, astro-ph/0206259.
3) Calchi Novati S., Iovane G, et al., Microlensing search towards M31, A&A 381, 848-861, 2002.
4) Capozziello S. and G.Iovane, On-Line Selection and Quasi-On-Line Analysis of Data using the Pixel Lensing Technique, Astrotech J., 1, 2, 1999.
5) Pratt W.K., Digital Image processing, Univ.South California, 1, 374, 1977.

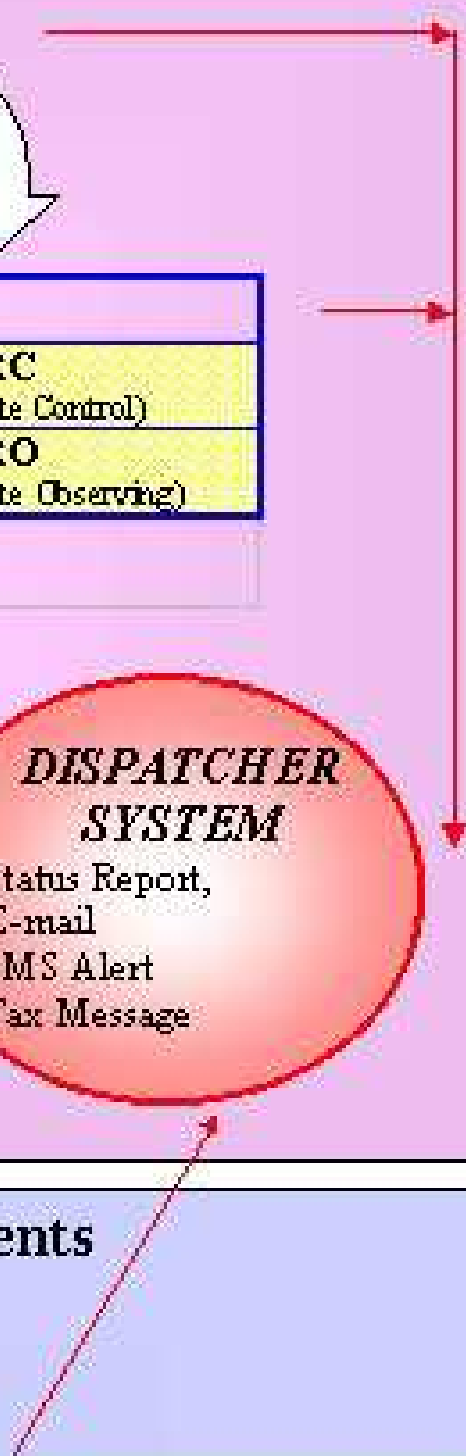

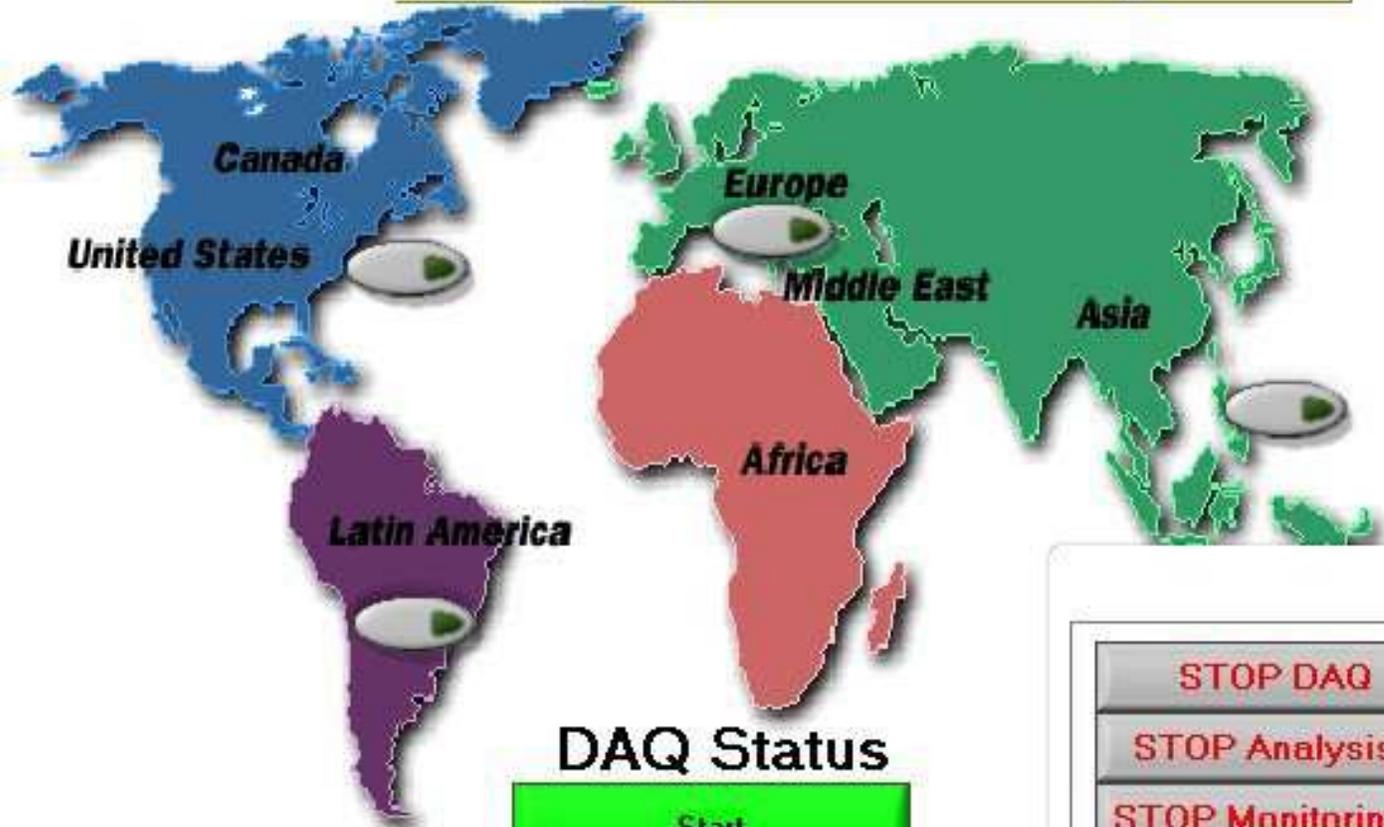

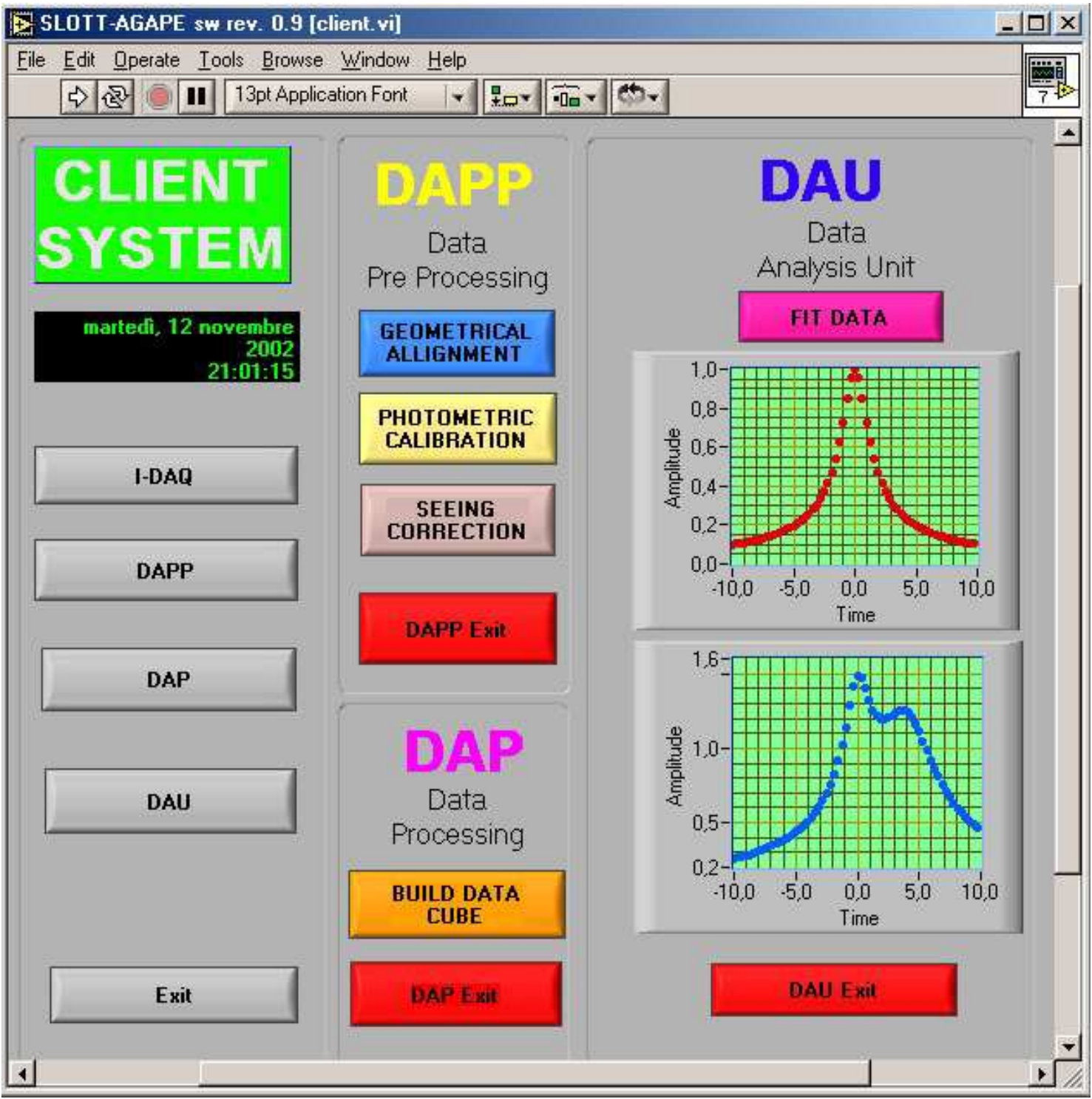

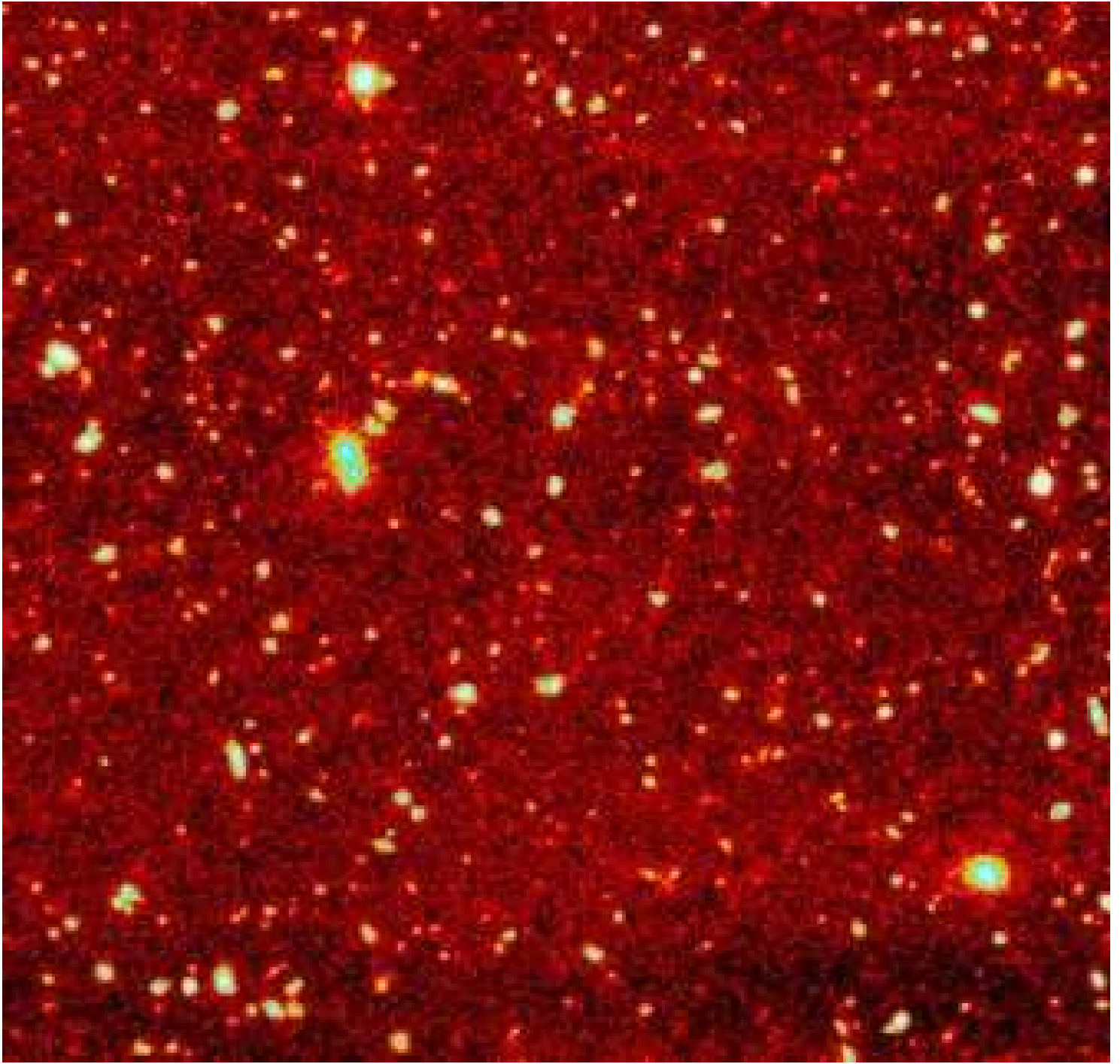

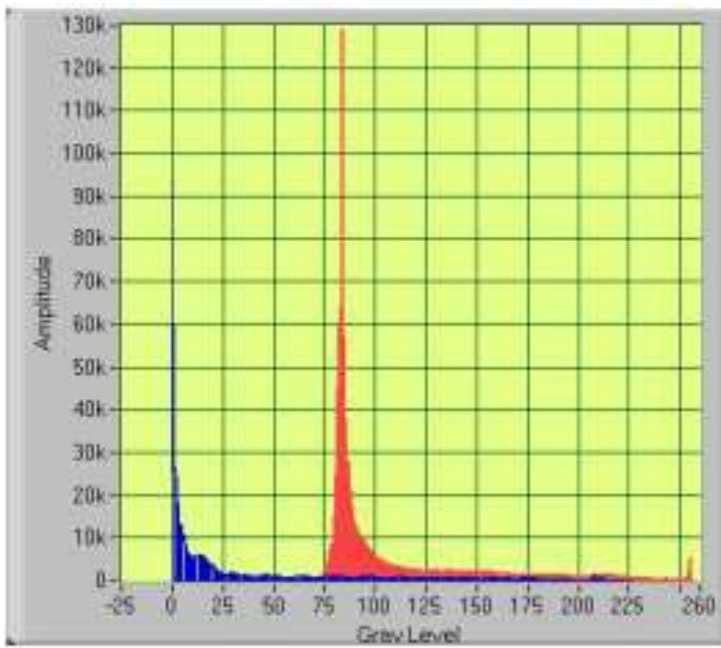
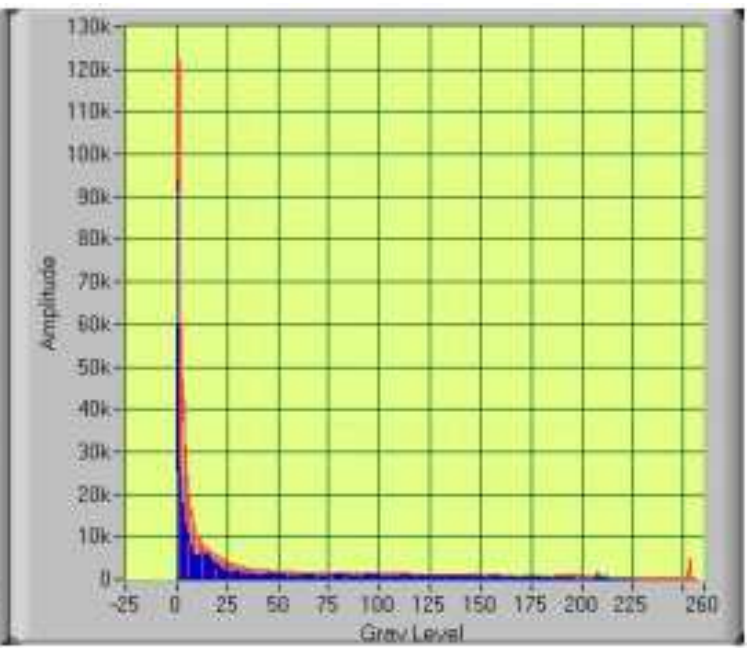

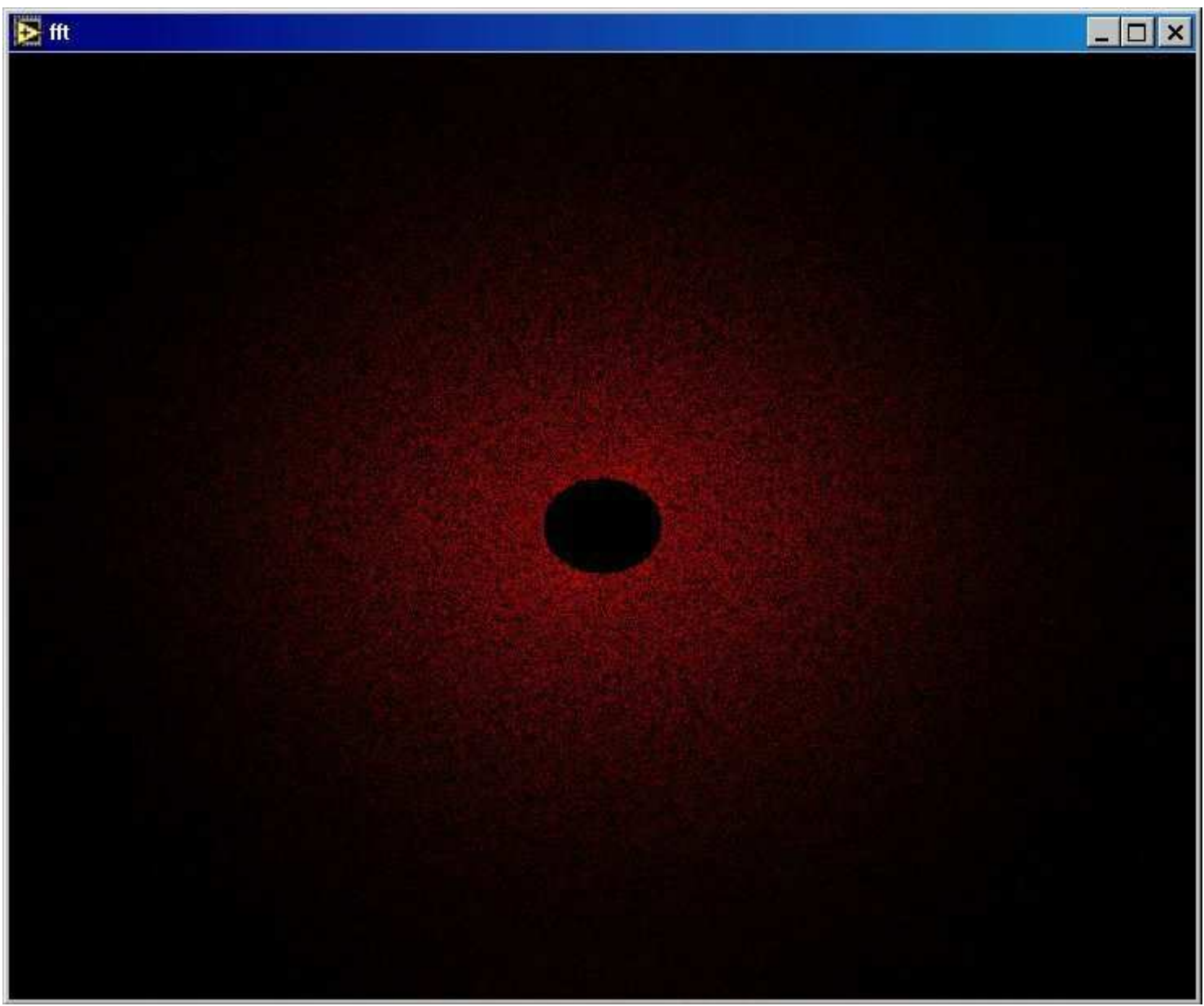

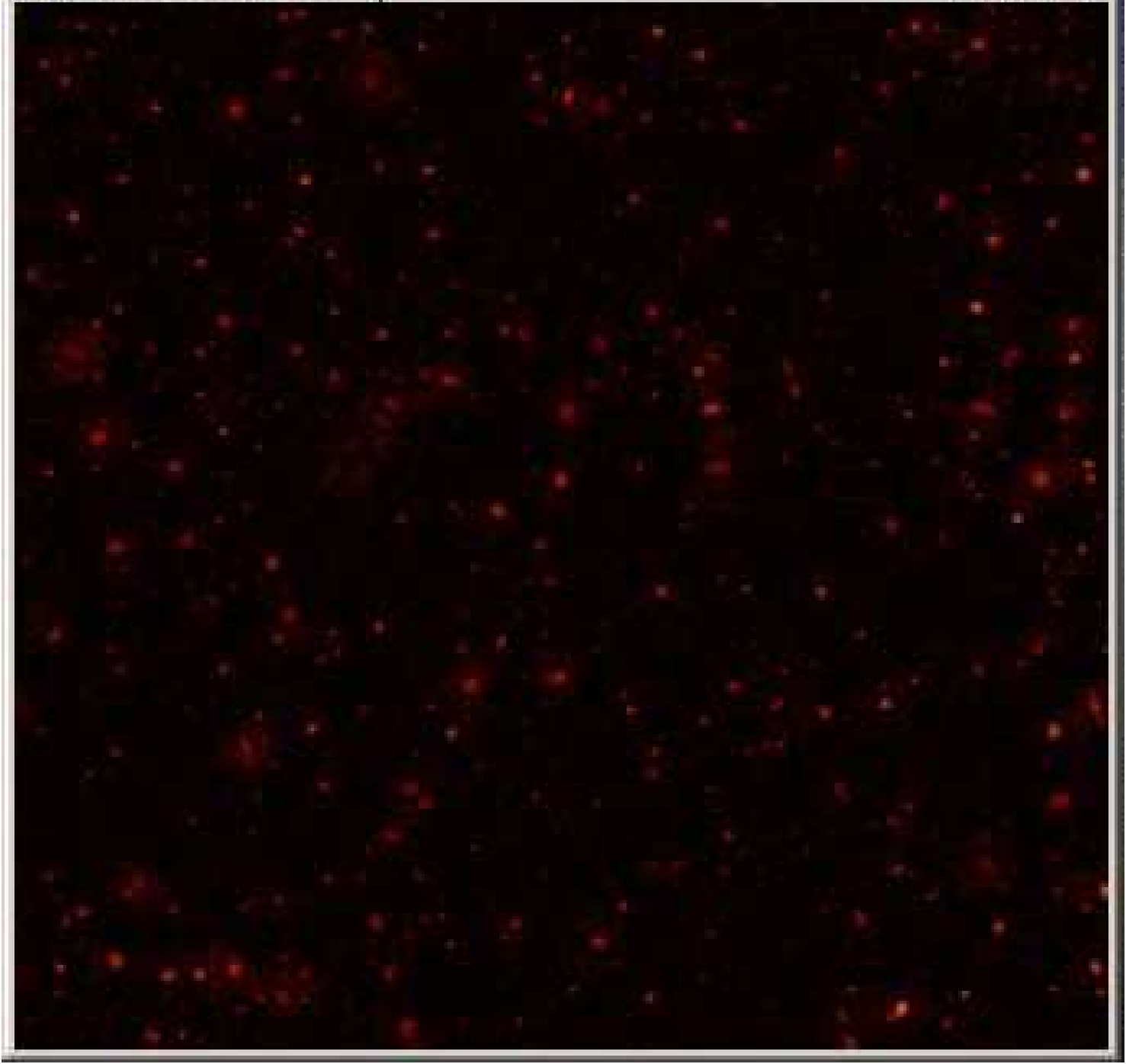

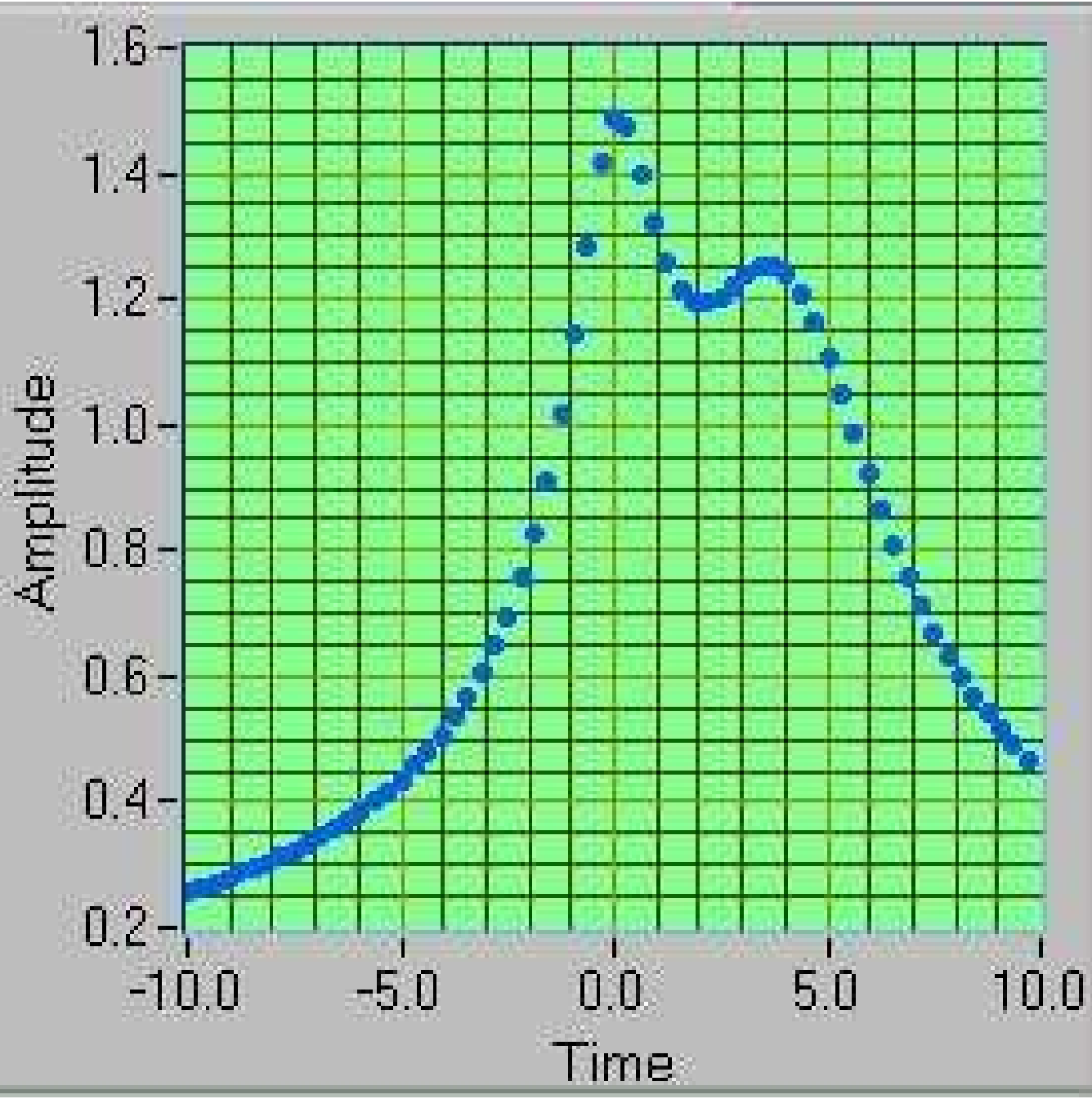